\documentclass[11pt]{article}

\pdfoutput=1

\usepackage{acl}

\usepackage{times}
\usepackage{latexsym}
\usepackage{tablefootnote}
\usepackage[T1]{fontenc}

\usepackage[utf8]{inputenc}
\usepackage{amsmath,amssymb}

\usepackage[noabbrev,capitalize]{cleveref}
\usepackage{booktabs}
\usepackage{multirow}
\usepackage{array}
\usepackage{microtype}

\usepackage{inconsolata}

\title{Tübingen-CL at SemEval-2024 Task 1:\\Ensemble Learning for Semantic Relatedness Estimation}

\author{Leixin Zhang \\
  University of T\"ubingen, Germany  \\
  \texttt{leixin.zh@gmail.com} \\\And
  Çağrı Çöltekin\\
  University of T\"ubingen, Germany  \\
  \texttt{cagri.coeltekin@uni-tuebingen.de} \\}

\begin{document}
\maketitle
\begin{abstract}

The paper introduces our system for SemEval-2024 Task 1, which aims to predict the relatedness of sentence pairs. Operating under the hypothesis that semantic relatedness is a broader concept that extends beyond mere similarity of sentences, our approach seeks to identify useful features for relatedness estimation. We employ an ensemble approach integrating various systems, including statistical textual features and outputs of deep learning models to predict relatedness scores. The findings suggest that semantic relatedness can be inferred from various sources and ensemble models outperform many individual systems in estimating semantic relatedness.

\end{abstract}

\section{Introduction}

Identifying semantic relatedness is a `related' task
to many well-studied tasks of semantic similarity.
According to \citet{abdalla-etal-2023-makes},
two sentences are considered similar if they are paraphrases or share
a relation of entailment.
Semantic relatedness,
however, is a broader concept than semantic similarity.
Two expressions are considered related if they share any semantic association.
For instance, `teacher' and `student' are related because
they frequently occur within the same context or domain.
Similarly, `tasty' and `unpalatable' are related,
as both terms are used to describe food, albeit with opposite meanings.

SemEval-2024 Task 1 \cite{semrel2024task} is designed to estimate
the relatedness of sentence pairs.
The task is based on a multilingual dataset of 14 languages
and offers supervised, unsupervised and cross-lingual tracks.
Our team participated in two tracks, and a subset of available languages:
Track A (supervised learning) for English,
and Track B (unsupervised learning) for English, Spanish, and Hindi.


We posit that semantic relatedness can be inferred from a multitude of sources and therefore propose an ensemble approach that integrates outcomes from diverse systems to estimate semantic relatedness. Our study explores features from textual statistical analysis, general large language models, word embedding models, and models trained on semantic labeled datasets, question-answering pairs, or title-passage pairs in estimating semantic relatedness, and we conducted ensemble experiments with these features.


\section{Related Work}

SemEval in previous years has introduced tasks focusing on semantic textual similarity to evaluate the degree of similarity between sentence pairs \cite{semeval-2012-sem,semeval-2013-joint-lexical,agirre-etal-2014-semeval,cer-etal-2017-semeval}. There tasks provided datasets with human labeled similarity scores, which have been extensively utilized for training sentence embedding models and conducting semantic evaluations \cite{wieting2015towards,cer2018universal, reimers-2020-multilingual-sentence-bert,feng-etal-2022-language}.

\subsection{Sentence Embeddings}

Word embedding models such as BERT \cite{devlin-etal-2019-bert}, GloVe \cite{pennington-etal-2014-glove}, RoBERTa \cite{liu2019roberta}, and Word2Vec \cite{mikolov2013distributed} are frequently employed to assess the semantic distance between words.  Sentence embeddings with a fixed length are often generated via mean/max pooling of word embeddings or employing \texttt{CLS} embedding in BERT. The semantic distances are commonly measured using the cosine similarity of embeddings of two expressions.

Siamese or triplet network architectures are frequently employed in sentence embedding training. For example, models such as Sentence-BERT \cite{reimers-gurevych-2019-sentence, reimers-2020-multilingual-sentence-bert} utilize a dual-encoder architecture with shared weights for predicting sentence relationships (e.g., semantic contradiction, entailment, or neutral labeling) or for similarity score prediction using regression objectives, e.g., the difference between human annotated similarity score (sim) of two sentences and the cosine of two sentence embeddings (${v}$ and ${u}$), illustrated in \cref{eq:1}.
\\
\begin{equation}
\label{eq:1}
 \mathcal{L} =  | \cos(v, u) - \text{sim} |
\end{equation}
\\
In triplet neural networks, an anchor sentence ($u$) can be trained along with a positive sample (a sentence with a similar meaning) and a negative sample (a sentence with a dissimilar meaning), with contrastive loss. InfoNCE (Noise-Contrastive Estimation) can be utilized as the objective function. A larger number of negative samples can also be integrated into neural networks through the application of InfoNCE, as demonstrated in \cref{eq:2}. Here, $v^+$ denotes positive samples. The negative sample size is denoted as $K$, and the total sample size (including one positive sample) as $K+1$. This approach is adopted by the Jina embedding model \cite{gunther-etal-2023-jina}, which is used in our ensemble system.\\ \\
\begin{equation}
\label{eq:2}
 \mathcal{L} =  -\mathbb{E} \left[\log  \frac{f(v^+,u)}{\sum_{i=1}^{K+1} f(v_i , u)}\right]
\end{equation}
\\

\subsection{Ensemble Learning}
In previous studies, ensemble learning presents several advantages. The ensemble approach can reduce the errors from individual models by amalgamating results from multiple sources or can make the system more robust.
In our study, using multiple pre-trained models can also save a substantial amount of computation  
while making use of information from the large data during pre-training. Previous research has demonstrated that ensemble learning can achieve remarkable success \cite{huang-etal-2023-ensemble,osika-etal-2018-second}.

In our study, we aim to integrate multiple deep learning models to assess semantic relatedness. When models are trained on diverse datasets with different architectures, they may produce varied predictions on semantic relatedness, and combining them may improve overall performance. 

We use sentence embeddings mainly from the following models. Sentence-BERT \cite{reimers-gurevych-2019-sentence} is trained on datasets involving SNLI (a collection of 570,000 sentence pairs) and MultiNLI (comprising 430,000 sentence pairs). The Jina Embedding model \cite{gunther-etal-2023-jina} utilizes 385 million sentence pairs and 927,000 triplets (comprising positive and negative samples of semantic similarity) after a filtering process. The T5 model is trained on approximately 7 TB of text data derived from Common Crawl, serving various text-to-text purposes \cite{raffel2020exploring,ni2021sentence}.



\section{Methodology}

In this study, we hypothesize that semantic relatedness covers a broader spectrum than semantic similarity in theory. Consequently, the integration of various systems and features should achieve superior results compared to individual systems.


\subsection{Supervised Learning} \label{feature}

For the supervised track\footnote{In the supervised track, we only participated English sub-task, in which relatively more training data was provided. For this reason, our analysis of supervised learning is specific to English.},
we first evaluated sub-systems in an unsupervised manner
and selected those with a higher Spearman's correlation
with human annotations for ensemble learning.
The selected results were then further fine-tuned using the training data (5,500 English sentence pairs labeled with relatedness scores provided by the shared task, \citealp{ousidhoum2024semrel2024}) to achieve closer alignment with human annotations.


In the following subsections, we present the features and systems utilized for ensemble learning.
The features can be classified into three categories: textual statistical features (\cref{statistical features}), word embedding models (\cref{word_embedding_models}), and sentence embedding models (\cref{sentence representation}).

\subsubsection{Textual Statistical Features} \label{statistical features}

Our analysis began with surface-level textual statistical features, including word overlap and the Levenshtein distance measurement at the character level. These scores were then normalized into ratios to estimate their correlation with human-annotated relatedness. Specifically, we considered the following features:
\begin{itemize}
    \item Character Distance Ratio: normalization of Levenshtein distance. Levenshtein distance (represented as ${Dist}$ in \cref{char-dist}) or edit distance is a string metric for measuring the difference or distance between two sequences at the character level. The character ratio we use in this study is defined as:
    
    \begin{equation} \label{char-dist}
    {\mathrm{\frac{len (Sent_1) + len(Sent_2) - Dist}{len (Sent_1) + len(Sent_2)}}}\\
    \end{equation}
    \item Word Overlap Ratio: 
    the count of overlapped words over the total word count in sentence pairs, expressed as:
    \begin{equation} \label{word-overlap}
    {\mathrm{ Ratio = \frac{|Words(A) \cap Words(B)|}{ |Words(A) \cup Words(B)|}}}
    \end{equation}
    \item  Content Word Overlap Ratio: the overlap ratio with content word considered only. Content words and functional words are distinguished by analyzing their part-of-speech (POS) using \texttt{SpaCy} python package. 
\end{itemize}
We found that the overlap ratio computed solely on content words shows a better correlation with the human judgment of relatedness (\cref{stat}). Furthermore, we tested the correlation of the word overlap ratio with the other two scores: Spearman's \textit{r} with content word overlap ratio is 0.77, and Spearman's \textit{r} with character distance ratio is 0.78. This suggests that the combination of two or more results may improve the relatedness estimation.

\begin{table}[!t]
    \centering
    \begin{tabular}{lc}
    \toprule
    \textbf{Statistic Features} & \textbf{Spearman \textit{r}}\\ \midrule
    {Char Distance Ratio} & 0.513 \\  
    {Word Overlap Ratio} & 0.593 \\  
    {Content Words Overlap Ratio} & 0.604\\ 
\bottomrule
\end{tabular}
\caption{Correlation between human-annotated relatedness scores with ratios of textual statistical features.}
 \label{stat}
\end{table}

\subsubsection{Word Embedding Models} \label{word_embedding_models}

In this subsection, we evaluate the performance of word embedding models' potential to estimate semantic relatedness. Sentence embeddings are represented as the mean of the word embeddings of all words in the sentence. We explored static word embeddings (GloVe and first layer BERT embeddings) and contextual word embeddings (the last layer of BERT embeddings) in relatedness estimation. The performance of the following variations is presented in \cref{word_embedding}:

 \begin{itemize}
     \item PCA transformation of embeddings. By using the PCA technique, we do not intend to reduce the dimension of the sentence embeddings, but transform sentence embeddings onto a new coordinate system such that the principal components capture the largest variation in the data. In practice, the maximum dimension that fits the dataset is adopted: \textit{min} (\textit{embedding\_length, sample\_size}).
    \item Content word embeddings: the average of word embeddings of content words only.
    \item Noun embeddings: the average of word embeddings for nouns only.
    \item Tree-Based word embeddings: the mean of embeddings of words that are at the top three levels of dependency trees,\footnote{We use SpaCy to parse sentences and select the root and dependents} namely the root (main predicate), direct dependents of the root, and dependents with the dependency distance of 2 from the root. 

 \end{itemize}

Our preliminary analysis offers the following insights for further ensemble learning:
\begin{enumerate}
    \item Excluding functional words (using content words only) can enhance the effectiveness of GloVe embedding. 
    \item Focusing on words closer to the sentence's `root' in terms of dependency distance did not yield better results. 
    \item  Contextualized BERT embeddings do not necessarily outperform uncontextualized embeddings in semantic relatedness estimation. 
    \item PCA-transformed embeddings show improved correlation with human annotation of relatedness.\footnote{Despite the better performance of PCA-transformed embeddings in Spearman's correlation when word embedding models are tested individually, it was not beneficial in later supervised training. Ultimately, GloVe$_{\text{Content}}$ word embedding was utilized in supervised and unsupervised ensemble learning for English.}
\end{enumerate}

\begin{table}[!t]
    \centering
    \begin{tabular}{lc}
    \toprule
   \textbf{Model} & \textbf{Spearman \textit{r} }\\ \midrule
    GloVe & 0.460 \\
    GloVe$_\text{PCA}$  & 0.533  \\ 
    GloVe$_\text{Content-words}$ & \textbf{0.554} \\
    GloVe$_\text{Tree-Based}$ & 0.249\\
    GloVe$_\text{Noun}$ & 0.430\\ \midrule
    BERT$_\text{LastLayer}$ & 0.399 \\
    BERT$_\text{LastLayer/PCA}$ & 0.446 \\
    BERT$_\text{FirstLayer}$  & 0.570 \\
    BERT$_\text{FirstLayer/PCA}$ & \textbf{0.593}\\
        \bottomrule
\end{tabular}
\caption{Spearman's correlation between human-annotated relatedness scores with the cosine similarity of average embeddings of all words, content words, all nouns or tree-based word selections within a sentence. PCA-transformed average embeddings of all words in a sentence are also presented.}
 \label{word_embedding}
\end{table}

\subsubsection{Models for Sentence Representations} \label{sentence representation}

For supervised learning, we also incorporate sentence representations from pre-trained language models into our ensemble system. This includes models known for their strong performance in sentence similarity tasks, involving Sentence-BERT (\texttt{mpnet-base}, \citealp{reimers-gurevych-2019-sentence}) and Jina Embedding (\texttt{jina-v1}, \citealp{gunther-etal-2023-jina}), as well as the general large language model, T5 encoder \cite{raffel-chen-2023-implicit, ni2021sentence}. Among all models tested in this study for English (refer to \cref{models}), T5 demonstrates the highest performance, achieving a Spearman's correlation of approximately 0.82 with human annotation.


\subsubsection{Ensemble Learning}


We explored two approaches for ensemble learning. The first approach operated directly on sentence representations from multiple models. This included concatenating sentence embeddings from various models and applying transformation (e.g., PCA transformation) in the embedding space to achieve a better correlation with human judgment. Our analysis indicates that while concatenation and transformation operations can slightly improve Spearman's correlation, they are not as effective as incorporating more statistical features into supervised fine-tuning. 

In the final system, we directly used the cosine similarity values from sentence embedding and word average embeddings as features (from models \texttt{mpnet-base}, \texttt{jina embedding}, \texttt{T5-base} and mean of content word embeddings from GloVe), along with textual statistic features (content word overlap ratio and character distance ratio) to estimate the relatedness of sentence pairs. These features are fed into Support Vector Machine (SVM) regression models (with RBF kernel) to predict human annotated relatedness.


\subsection{Unsupervised Ensemble}

In the unsupervised track, without utilizing labeled datasets for sentence similarity or relatedness and without employing models pre-trained on labeled datasets, we aim to evaluate whether models trained on other types of datasets intended for different purposes could generate representations suitable for estimating semantic relatedness.

In addition, we investigated whether integrating additional features, such as the cosine distance of average word embeddings and word overlap ratios, could enhance performance. We calculated the arithmetic mean of the cosine distances and ratios from textual statistics as the relatedness prediction of sentence pairs. Various feature combinations are tested with the provided validation dataset. 
 
 For the unsupervised task of English, we utilized two models to generate sentence representations: a model designed for semantic search (\texttt{multi-qa-MiniLM-L6-cos-v1}, \citealp{reimers-gurevych-2019-sentence}), trained on 215 million question-answer pairs; and e5 (\texttt{e5-base-unsupervised}, \citealp{wang2022text}),\footnote{The e5 monolingual model is exclusively used for English, not for the other two languages: Spanish and Hindi} trained on question-answer pairs, post-comment pairs, and title-passage pairs. These models were further refined with an unsupervised transformation (PCA). Additionally, we incorporated two other features: PCA-transformed GloVe embeddings (average of content word embeddings within a sentence) and content word overlap ratios into the unsupervised ensemble system.
 
For the unsupervised tasks in Spanish and Hindi, we used a similar method for predicting relatedness, combining features involving the cosine distance of \texttt{multi-qa-MiniLM} model representations, word embedding model and word overlap ratios. For word embeddings, we employed multilingual BERT (\texttt{bert-base-multilingual-uncased}), utilizing both the first-layer (uncontextualized) and last-layer (contextualized) embeddings for relatedness estimation.


\section{Results and Analysis} 

The shared task evaluates the participating systems based
on Spearman's correlation (\textit{r}) between the human-annotated scores,
which ranges from 0 to 1. 
In \cref{models}, we compare the correlation scores
for our systems and other popular models on the official test set.

\begin{table}[!t]
    \centering
    \begin{tabular}{lccc}
    \toprule
     Models& English & Spanish & Hindi \\
     \midrule
     Lexical Overlap&0.741&0.661&0.587\\
     mBERT$_\text{\_Ave}$&0.640&0.655&0.566 \\
     mpnet-base\tablefootnote{Table 3 shows \texttt{all-mpnet-base-v2} result for English and \texttt{paraphrase-multilingual-mpnet-base-v2} model results for Spanish and Hindi, model details: https://www.sbert.net/docs/pretrained\_models.html} & 0.809 &0.590& \textbf{0.746}\\
        T5 (base) & 0.825 &- &-\\
        LaBSE\tablefootnote{\citealp{feng-etal-2022-language}
} &0.818 & 0.651 & 0.709\\
        multi-qa-Mini & 0.793 & 0.638 & 0.466 \\
       Ensemble$_\text{Sup}$ &\textbf{0.850}&-&-\\
       Ensemble$_\text{Unsup}$ & \textbf{0.837} & \textbf{0.705} & 0.649 \\
\bottomrule
\end{tabular}
\caption{Spearman correlation between human-annotated relatedness scores and system predicted scores on the test dataset.}
 \label{models}
\end{table}

Results presented in \cref{models}  suggest that the ensemble approach generally outperforms single models. Specifically, the ensemble system trained with true labels, for the supervised English task, achieved the best result among all listed systems, with an improvement in Spearman's correlation of 0.025 compared to the T5 base model.

The ensemble approach for English and Spanish unsupervised tasks also achieved relatively high scores, despite the absence of similarity or relatedness scores in learning. It suggests that semantic relatedness can be estimated without necessarily relying on human-annotated scores of semantic similarity or semantic relatedness. Other sources like question-answering pairs or statistical features of texts also play a role in relatedness estimation. Thus, the ensemble of statistical text features, word embedding models, and models trained on question-answer pairs can achieve good results.

Although the results for Hindi did not match the superior outcomes of other supervised models, such as \texttt{mpnet-base} and \texttt{LaBSE}, which were trained with semantic labels or similarity scores, the ensemble system's performance still surpasses that of the multilingual BERT embedding model and the \texttt{multi-qa} model, both of which were utilized for ensemble learning as base models.




\subsection{Biased Performance}
We also observe that the unsupervised results for Hindi are not comparable with those from Spanish and English though with the same ensemble approach. This discrepancy stems from the suboptimal performance of the sub-models used in the unsupervised ensemble. For example, the multi-qa-MiniLM model utilized for Hindi only achieves a correlation of 0.466, and the multilingual BERT for Hindi is also less effective compared to the other two languages. 

Apart from Hindi, we also applied the same ensemble method to other non-Indo-European languages in the unsupervised track, yet the results scarcely surpassed 0.60 for the validation dataset, so results of other languages were ultimately not submitted.

The results indicate that some multilingual models are biased towards English and Indo-European languages, and perform less effectively for other languages. This bias may be attributed to imbalanced data during the models' pre-training phase.



\section{Conclusion}

Our system employs an ensemble approach to estimate semantic relatedness, integrating results from multiple systems: textual statistical features, word embedding models, and sentence representation models. Our findings suggest that semantic relatedness can be deduced from a variety of sources. Although some features (e.g., lexical overlap ratio) may not perform as strongly as models specifically designed to obtain sentence representations, the results demonstrate that these features, when used in a combined manner, can outperform many individual systems and collaboratively achieve a better correlation with human judgment on semantic relatedness.

\section{Limitation and Future Work}
Constrained by the size of the training data and the availability of pre-trained language models, it is regrettable that we did not offer insights into other Asian and African languages. In future research, studies on low-resource languages will be valuable, including tasks such as data collection, annotation, and pre-training models tailored to these languages.

\section*{Acknowledgements}
We are very grateful for the assistance and discussions provided by Leander Girrbach and Milan Straka.

\bibliography{anthology,custom}

\appendix



\end{document}